\documentclass[letterpaper]{article} 
\usepackage{aaai24}
\usepackage{times}  
\usepackage{helvet}  
\usepackage{courier}  
\usepackage[hyphens]{url}  
\usepackage{graphicx} 
\usepackage{natbib}  
\usepackage{caption} 
\usepackage{algorithm}
\usepackage{algorithmic}

\usepackage{newfloat}
\usepackage{listings}
\DeclareCaptionStyle{ruled}{labelfont=normalfont,labelsep=colon,strut=off} 
\floatstyle{ruled}
\newfloat{listing}{tb}{lst}{}
\floatname{listing}{Listing}
\newcommand{\sysname}{YTCommentQA}
\usepackage{booktabs}
\usepackage{array} 
\usepackage{tabularx}
\usepackage{multirow}
\usepackage{makecell}
\usepackage{graphicx}
\usepackage{xcolor, colortbl}
\title{\sysname{}: Video Question Answerability in Instructional Videos}
\author{
    Saelyne Yang\textsuperscript{\rm1}\thanks{Work done at LG AI Research},
    Sunghyun Park\textsuperscript{\rm 2},
    Yunseok Jang\textsuperscript{\rm3}\footnotemark[1],
    Moontae Lee\textsuperscript{\rm 2,4}
}
\affiliations{
    \textsuperscript{\rm 1}KAIST,\\
    \textsuperscript{\rm 2}LG AI Research,\\
    \textsuperscript{\rm 3}University of Michigan,\\
    \textsuperscript{\rm 4}University of Illinois Chicago\\
    saelyne@kaist.ac.kr,
    \{sunghyun.park, moontae.lee\}@lgresearch.ai,
    yunseokj@umich.edu

}

\usepackage{bibentry}

\begin{document}

\maketitle
\begin{abstract}
Instructional videos provide detailed how-to guides for various tasks, with viewers often posing questions regarding the content. Addressing these questions is vital for comprehending the content, yet receiving immediate answers is difficult. While numerous computational models have been developed for Video Question Answering (Video QA) tasks, they are primarily trained on questions generated based on video content, aiming to produce answers from within the content. However, in real-world situations, users may pose questions that go beyond the video's informational boundaries, highlighting the necessity to determine if a video can provide the answer. Discerning whether a question can be answered by video content is challenging due to the multi-modal nature of videos, where visual and verbal information are intertwined. To bridge this gap, we present the \sysname{} dataset, which contains naturally-generated questions from YouTube, categorized by their answerability and required modality to answer -- visual, script, or both. Experiments with answerability classification tasks demonstrate the complexity of \sysname{} and emphasize the need to comprehend the combined role of visual and script information in video reasoning.
The dataset is available at {\url{https://github.com/lgresearch/YTCommentQA}}.
\end{abstract}
\section{Introduction}

\begin{figure}[t]
\centering
\includegraphics[width=\linewidth]{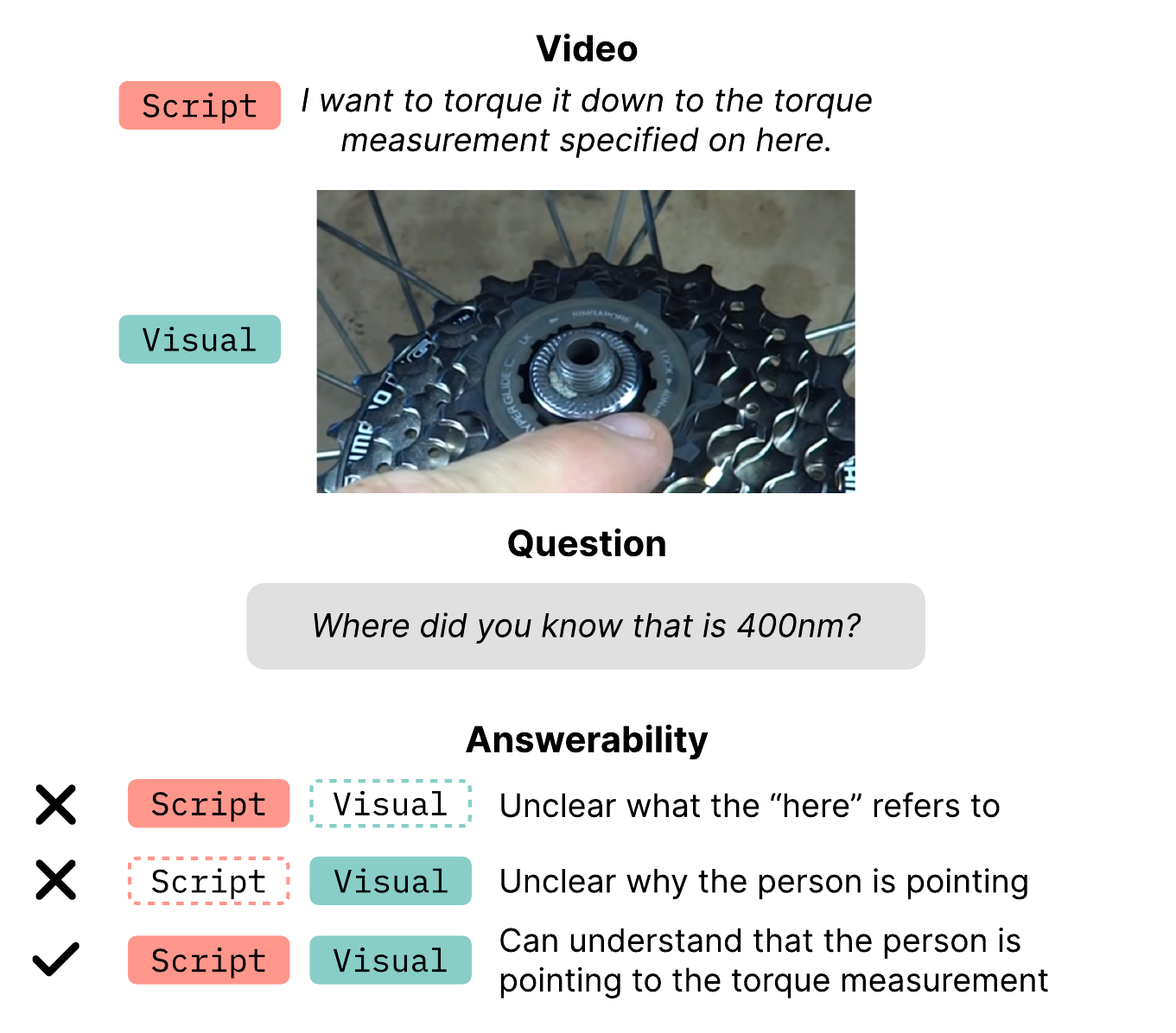}
\caption{A question on video can be either (1) unanswerable by video, (2) answerable by visual, (3) answerable by script, or (4) answerable when both visual and script are present. The figure shows an example of (4), where the question is answerable with the understanding of both visual and script.}
\label{figure:concept}
\end{figure}

Instructional videos present a procedural guide on how to perform a task through visual demonstrations and verbal explanations \cite{Miech2019HowTo100MLA}. As people watch and follow the videos, they naturally have questions regarding the video content, such as clarifying a certain part or adapting the instructional content to their own situations \cite{Madden2013ACS, scheme-coding}. Providing appropriate answers to these questions is crucial for people to comprehend and follow instructions effectively. However, addressing every question promptly can pose challenges for video creators. 

At the same time, extensive research has been conducted in the domain of Video Question Answering (Video QA), which aims to provide answers to questions about video content \cite{Colas2019TutorialVQAQA, zhao2021photoshop, Li2020HeroHE, Yang2020JustAL}. 
However, the majority of models are trained on datasets labeled by crowd workers or domain experts who were instructed to create both questions and answers based on the video content \cite{Colas2019TutorialVQAQA, zhao2021photoshop, Li2020HeroHE}, or those generated with automated methods \cite{Yang2020JustAL}. 
Yet, questions from real-world users can include queries that go beyond the scope of what can be answered within the video while questions still remain relevant to the content. 
These queries may necessitate additional knowledge such as domain or creator-specific knowledge \cite{zhao2021photoshop}. 
To avoid generating false information, it is essential to determine whether a given question can be answered within the video. 
However, to our knowledge, there has been limited prior research on video answerability.

\begin{table*}[t]
\centering
\setlength{\tabcolsep}{10pt}
\begin{tabular}{llll}
\toprule
\textbf{Dataset} & \textbf{Domain} & \textbf{Question Collection} & \textbf{Question Answerability}\\ \midrule
\textbf{TutorialVQA} & Screencast Tutorial & Crowd workers & Answerable in video \\ 
\textbf{PsTuts-VQA} & Screencast Tutorial & Domain experts & \makecell[tl]{Answerable \\ with domain knowledge base} \\ 
\textbf{HowToVQA69M} & General How-to & Automated method & Answerable in video \\ 
\textbf{iVQA} & General How-to & Crowd workers & Answerable in video \\ 
\textbf{How2QA} & General How-to & Crowd workers & Answerable in video \\ \midrule
\textbf{\sysname (Ours)} & General How-to & Real-world users & \makecell[tl]{ Answerable \& Unanswerable \\ (Answerable contains required modality)} \\
\bottomrule
\end{tabular}
\caption{Comparison of existing instructional video QA datasets and \sysname{}}
\label{table:dataset_comparison}
\end{table*}


The Video Answerability task poses significant challenges due to the multi-modal nature of videos, where information is conveyed through both visual and verbal channels.
A simplistic approach to determine answerability is to extract frames and transcripts from videos and apply existing techniques for text answerability~\cite{rajpurkar-etal-2018-know} and image answerability~\cite{Gurari2018VizWizGC}. 
However, relying solely on these approaches is insufficient because in videos, visual and verbal information complement each other~\cite{Huang2018FindingW}, and crucially, some questions can be answered only when both modalities are considered together (See Figure \ref{figure:concept}).
Also, image answerability alone may not suffice since there are instances where a single frame does not provide an answer, but a series of frames can, such as repetition count~\cite{jang-IJCV-2019}.

This paper presents \sysname{}, a comprehensive dataset containing naturally generated questions from real-world users in YouTube instructional videos. The dataset includes questions that naturally arise from real-world users in the comment sections of YouTube, demonstrating greater diversity and length than questions found in existing Video QA datasets. It also indicates whether the question is answerable within the video and, if so, provides the associated answer segments. The answerability information categorizes questions into four groups depending on the required modality to answer the questions: (1) Unanswerable, (2) Answerable with visuals, (3) Answerable with the script, and (4) Answerable only when both script and visuals are provided.

We design two tasks for the \sysname{} dataset. The Segment Answerability Classification asks the model to predict the answerability of questions given the evidence segment, while the Video Answerability Classification task asks the model to predict answerability along with the required modality given the entire video. The results demonstrate the challenging nature of \sysname{} in predicting the answerability of questions and its required modality. Our dataset highlights the need to understand the complementary role of visual and script information in video question answering. 

\section{Related Work}
\subsection{Video Question Answering}

\begin{figure*}[t]
\centering
\includegraphics[width=0.9\linewidth]{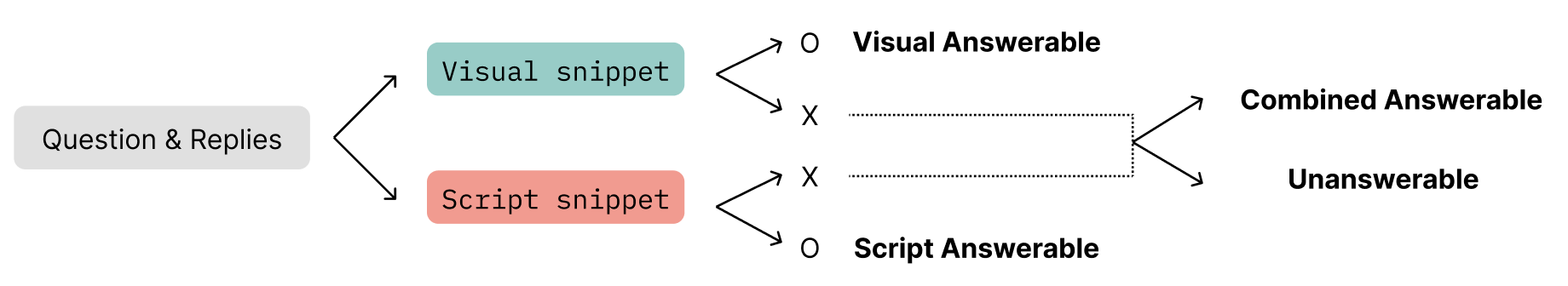}
\caption{Annotation workflow for the video question answerability. Once an annotator identifies that the \texttt{timestamp} used in a reply to a given question suggests an answer in the video, they are provided with visual and script snippets centerd around the timestamp. For questions that could not be answered using visual or script snippets, the annotators are asked whether both were necessary to answer or if the question was unanswerable altogether.}
\label{figure:workflow}
\end{figure*}

To enhance the comprehension of videos through question and answering, numerous computational approaches have been devised~\cite{yang2020bertVideoQA, Ye2017VideoQA, yang2022frozenbilm} and a range of benchmark datasets have been proposed. As succinctly outlined by Zhong, Xiao, and Ji et al., Video QA datasets can be classified into three principal categories based on the information they engage: Plain, Multimodal, and Knowledge-based Video QA~\cite{zhong-etal-2022-video}: (1) Plain Video QA delves into the visual content of videos, encompassing understanding everyday activities ~\cite{yu2019activityqa, castro-etal-2022-in-the-wild}. Given the intrinsic complexity arising from the spatial and temporal nature of videos, several datasets focus on the spatial, temporal, and causal reasoning within videos ~\cite{GrundeMcLaughlin2021AGQA, jang-IJCV-2019, xiao2021next}. (2) Multimodal Video QA extends its reach beyond visuals by incorporating supplementary information from videos, such as subtitles. This multifaceted approach empowers Video QA with a profound understanding encompassing both visual and verbal components~\cite{castro-etal-2020-lifeqa, lei-etal-2018-tvqa, Tapaswi2015MovieQAUS}. (3) Lastly, knowledge-based Video QA draws upon external sources beyond the video itself. These sources might encompass, for instance, commonsense knowledge gleaned from a drama series~\cite{Garca2019KnowITVA} or software application-specific knowledge~\cite{zhao2021photoshop}. This Multimodal Video QA highlights the crucial role played by visual and verbal elements, and the knowledge-based Video QA highlights that there are questions that cannot be answered by the video that users are still interested in. We aim to foster this line of research by detecting which question needs additional knowledge in the first place by investigating the nuanced role played by each modality.

\subsection{Instructional Video Question Answering}
Focusing on instructional videos, TutorialVQA~\cite{Colas2019TutorialVQAQA} and PsTuts-VQA~\cite{zhao2021photoshop} have focused on screencast videos, aiming to cultivate a comprehensive understanding of software tutorial videos. Expanding the scope, HowToVQA69M~\cite{Yang2020JustAL}, iVQA~\cite{Yang2020JustAL} and How2QA~\cite{Li2020HeroHE} have harnessed questions extracted from the extensive HowTo100M dataset, an expansive repository of how-to videos spanning 12 distinct genres~\cite{Miech2019HowTo100MLA}. However, their questions are artificially generated and might be far from what real-world users would have who follow the tutorial content. Some have gathered questions manually, presenting answer segments to crowdsourced workers and soliciting question generation~\cite{Colas2019TutorialVQAQA,Li2020HeroHE, Yang2020JustAL}. Others have relied on domain experts for crafting question-answer pairs~\cite{zhao2021photoshop} or automated techniques~\cite{Yang2020JustAL}. Our dataset not only tackles the video question answerability problem but also is enriched by the incorporation of naturally generated questions, thus infusing a more human-like and authentic dimension to the landscape of Video QA.

\subsection{Text and Image Question Answerability}
Identifying whether a question can be answered or not from a given context is important for reliable and interpretable answer generation~\cite{Hu2018ReadV, Nishida2021TowardsIA}. In the domain of machine reading comprehension with text-based context, numerous studies have contributed to understanding unanswerable questions. SQuAD 2.0~\cite{rajpurkar-etal-2018-know} introduces relevant unanswerable questions, challenging models to discern whether a question can be answered or not. Several other question-answering datasets have also incorporated unanswerable questions to improve reading comprehension~\cite{Choi2018QuACQA, Reddy2018CoQAAC, Campos2016MSMA, trischler-etal-2017-newsqa}. Additionally, data augmentation techniques, like generating relevant unanswerable questions~\cite{Zhu2019LearningTA}, have been introduced to enhance this line of research.


Question answerability has also been explored in the domain of visual question answering. A line of work augmented existing Visual Question Answering datasets with irrelevant questions to adequately handle unanswerable cases~\cite{Ray2016QuestionRI, mahendru-etal-2017-promise, toor2017c2vqa}. Additionally, VizWiz~\cite{Gurari2018VizWizGC} collected naturally occurring unanswerable questions from associated images taken by blind individuals. These images feature inadequate image quality or content, presenting real-world challenges in VQA. To predict answerability in VQA, several transformer-based models have been proposed~\cite{Le2021VisionAT, NguyenTran2022RVTTransformerRA}

In light of these contributions to answerability in both text and image contexts, to our knowledge, there is no existing work on answerability in videos. To address this gap and further enhance question-answering systems, we propose a novel dataset for video question answerability that requires multi-modal reasoning skills. We believe that exploring answerability in videos can advance the understanding of question answerability in more diverse and complex scenarios.
\section{\sysname{}: Dataset Collection}

We introduce \sysname{}, a video question answerability dataset. 
It consists of 2,332 questions asked by real-world users on 2,004 YouTube videos and their associated answerability information. 
We describe our dataset collection procedure below.

\subsection{Video Collection}
We selected videos from the YT-Temporal-180M dataset~\cite{Zellers2021MERLOTMN}, which encompasses a wide range of domains and topics from public YouTube videos. We specifically focused on instructional videos present in the dataset, which overlaps with the HowTo100M~\cite{Miech2019HowTo100MLA} dataset. Then, we further refined our selection by filtering videos falling within the ``Howto \& Style" category on YouTube, resulting in 245,354 videos.\footnote{We filtered the videos according to the category because there were non-how-to videos in the initial selection, such as music videos.} We obtained the transcripts of the videos and restored punctuations in the full transcript using an off-the-shelf BERT-based model~\cite{bert-huggingface}, where completed sentences were segmented by full stops.  

\subsection{Question Collection}

\begin{figure}[t]
\centering
\includegraphics[width=\linewidth]{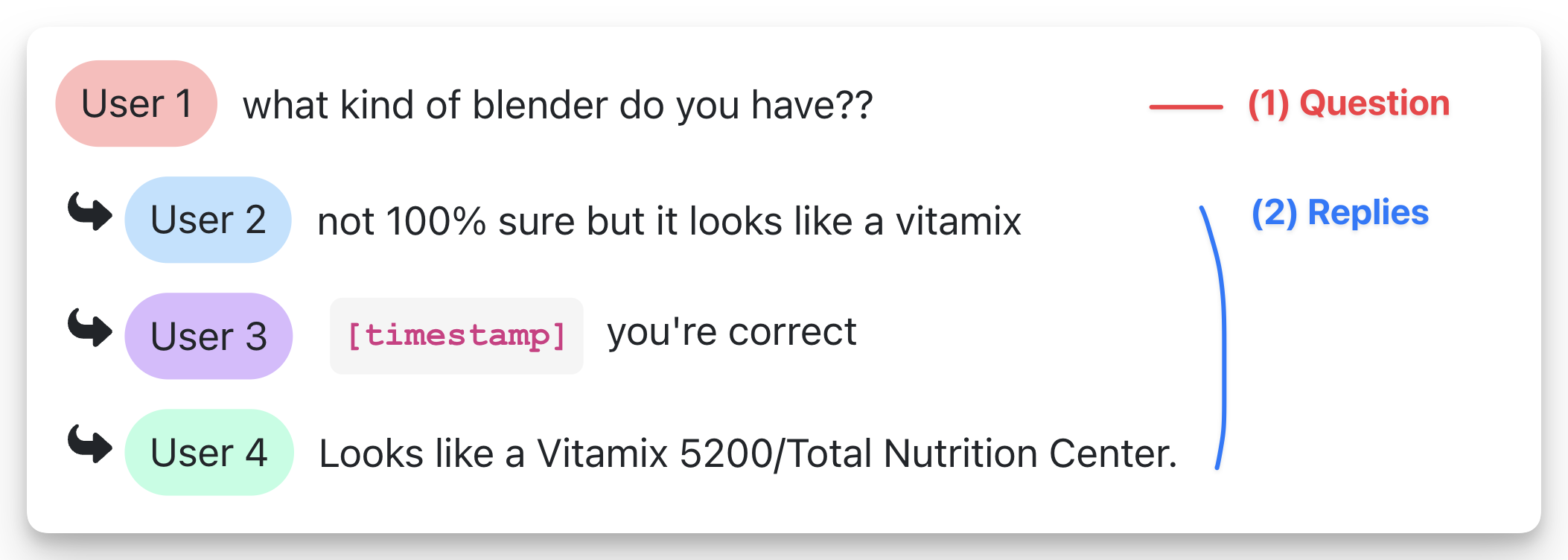}
\caption{Example question and its replies that contain a timestamp.}
\label{figure:conversation}
\end{figure}

To gather a diverse range of questions relevant to the video content, we collected queries generated by actual users. To achieve this, we crawled the comment section on YouTube for each video utilizing the YouTube Data API~\cite{ytdata-api}. We employed a language detection library~\cite{lang-detect} to filter out non-English comments. After that, we utilized a question detection module~\cite{question-detect} to extract and retain only the comments that contained questions. 

For the collection of real-world questions and answers that pertain to the video content, we devised the following process. First, we selected questions that had one or more replies and a timestamp (in the format of \texttt{dd:dd}) within the replies. The inclusion of timestamps in comments is a common practice where users refer to specific points in the video~\cite{yarmand2021can}. We anticipated that timestamps used in replies would likely point to video relevant to the answers, streamlining the process for crowd workers to identify the answerability of questions. Next, we retained only those questions where the responses involved at least one user other than the original commenter. We excluded questions that had conversations involving more than five users to prevent excessively lengthy threads. This resulted in 3,260 questions from 2,912 videos, which were later narrowed down to 2,332 after another filtering in the Answerability Annotation phase (Section~\ref{sec:answerability_annotation}). In Appendix~\ref{sec:compare}, we provide a comparison between  \sysname{} questions and general questions that include questions that do not have replies with a timestamp.


\begin{table}[t]
\centering
\setlength{\tabcolsep}{17pt}
\begin{tabular}{ll}
\toprule
Videos & 2004 \\
Duration (seconds) &  524.32±262.8 \\ \midrule
Questions & 2332 \\
Questions per video & 1.16±0.62 \\
Question length (\# words) & 17.2±15.06 \\ \midrule
Evidence length (seconds) &  27.32±17.1 \\
Evidence length (\# words) & 67.34±37.13 \\ \midrule
(1) Unanswerable & 326 (13.98\%) \\
(2) Visual Answerable & 1427 (61.19\%) \\
(3) Script Answerable & 1155 (49.53\%) \\
(4) Combined Answerable & 174 (7.46\%) \\
\bottomrule
\end{tabular}
\caption{Data Statistics for \sysname{}. Note that (2) and (3) have overlapping questions.}
\label{table:stats}
\end{table}

\subsection{Answerability Annotation} \label{sec:answerability_annotation}

To annotate the answerability for each question, we developed an annotation system (Appendix~\ref{sec:system}) and recruited annotators through Prolific~\cite{prolific}. 
We designed a two-step process which we describe below.

First, since not all strings in timestamp format (\texttt{dd:dd}) necessarily means that the timestamp contains the answers, annotators were presented with the entire conversation (i.e., a question and its replies. See Figure~\ref{figure:conversation} for an example) and asked if it \textit{implied} that the answer to the question could be found around the timestamp. If the response was positive, we went to the second step, where we displayed a video snippet without audio source near the timestamp and queried whether the visual snippet provided the answer. Participants had two options: ``yes" or ``no". We then showed the corresponding transcript and inquired if the answer was available in the text. Participants had three response options: ``yes", ``no", or ``requires both video and transcripts." Participants could extend the visual or transcript by up to three more sentences (corresponding video segments for visual snippets) if the snippet ended in the middle of the answer. If the response to the first stage was negative, we asked participants to indicate the alternate purpose of the timestamps being used. We did not include such questions in our dataset since the identified text as a timestamp did not relate to a potential answer segment.

In the entire process, to ensure quality control, we assigned two participants to each question, and in cases of disagreement, we sought one more participant's response until a single answer received at least two votes. Responses with less than 2/3 overlap with others were discarded. Participants were compensated with 0.17 GBP on average for annotating one question.

In summary, the annotated answerability information includes the following categories: (1) Unanswerable, (2) Answerable with visual, (3) Answerable with the script, (4) Answerable when both script and visual are provided. Note that (2) and (3) may co-exist. The complete annotation workflow is depicted in Figure~\ref{figure:workflow}.
Ethical considerations are discussed in Appendix~\ref{sec:ethics}.

\begin{figure}[t]
\centering
\includegraphics[width=0.9\linewidth]{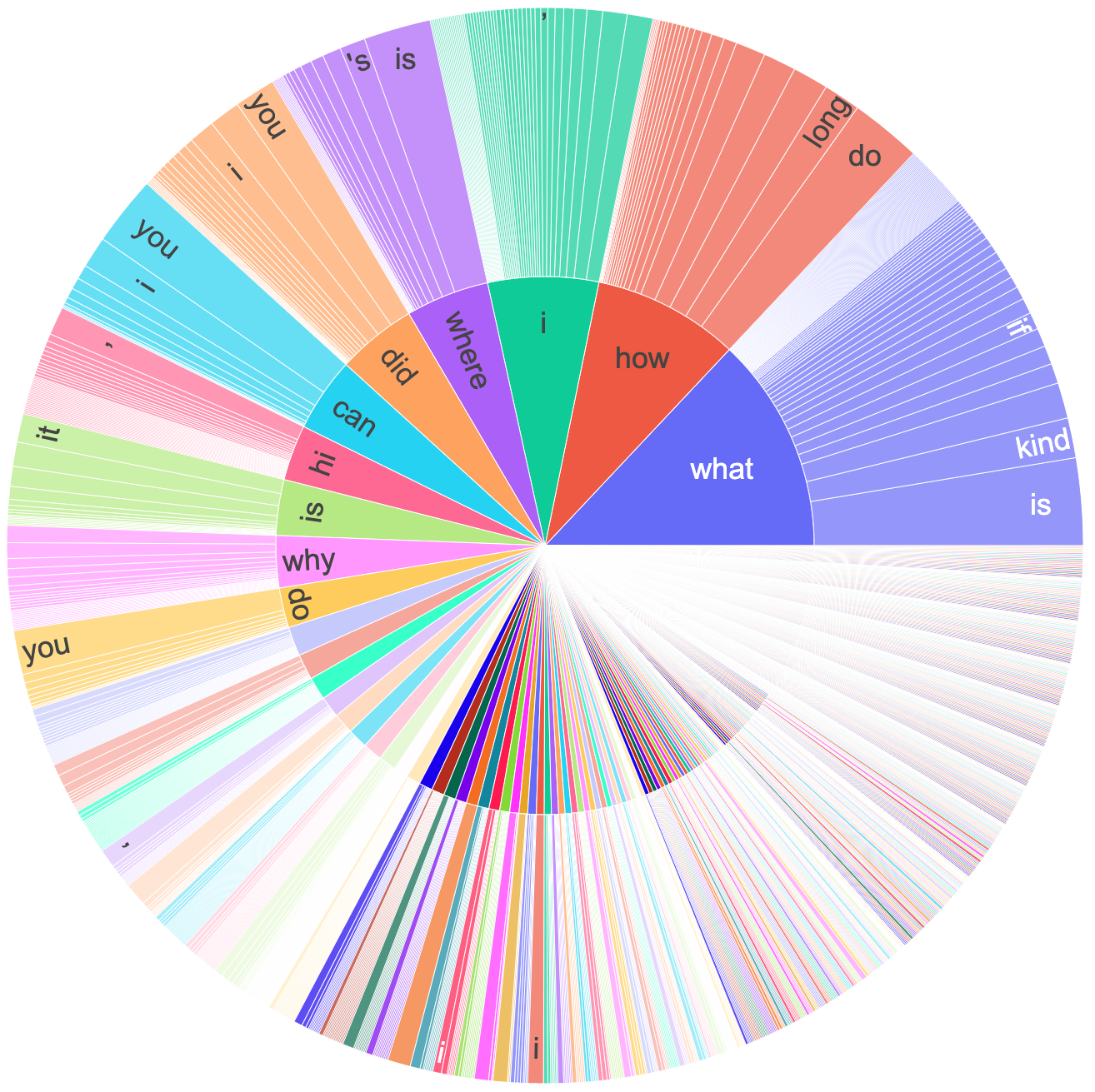}
\caption{Distribution of the first two words for questions in \sysname{}, which shows the diversity of the collected questions. The sequence of words begins from the center and extends outward. Words with small font sizes are omitted.}
\label{figure:sunburst}
\end{figure}

\section{\sysname{}: Dataset Analysis}

\subsection{Analysis of Questions}

\begin{table*}[hbt!]
\centering
\begin{tabular}{>{\centering\arraybackslash}m{2cm}>{\centering\arraybackslash}m{4.5cm}>{\centering\arraybackslash}m{4.5cm}>{\centering\arraybackslash}m{4.5cm}}
\toprule
\textbf{Answerability} & \textbf{Visual} & \textbf{Script} & \textbf{Question} \\ \hline
\multirow{2}{*}{\parbox{2cm}{\vspace{1.2cm} \centering Visual\\Answerable}}   & {\includegraphics[width=0.25\textwidth]{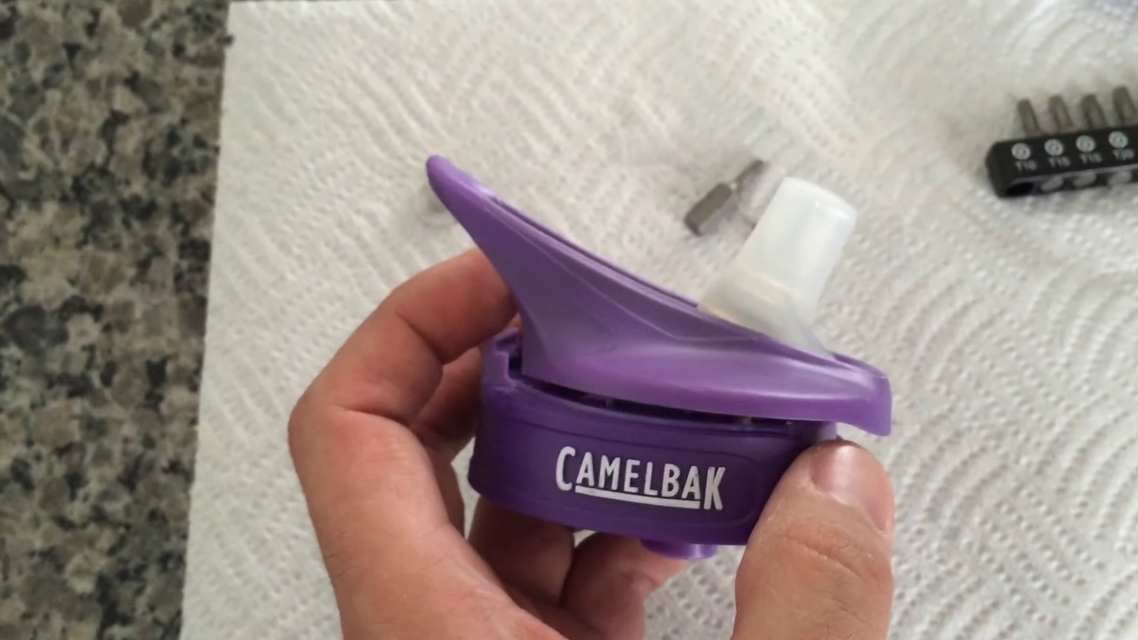}} & {you can see the damage on my bottle and the the top part comes off to the front rather than up.} & {i can't figure out how to take the top part off of the bottom in order to get the air vent back in. can you help me? thanks!!} \\ \cline{2-4}
& {\includegraphics[width=0.25\textwidth]{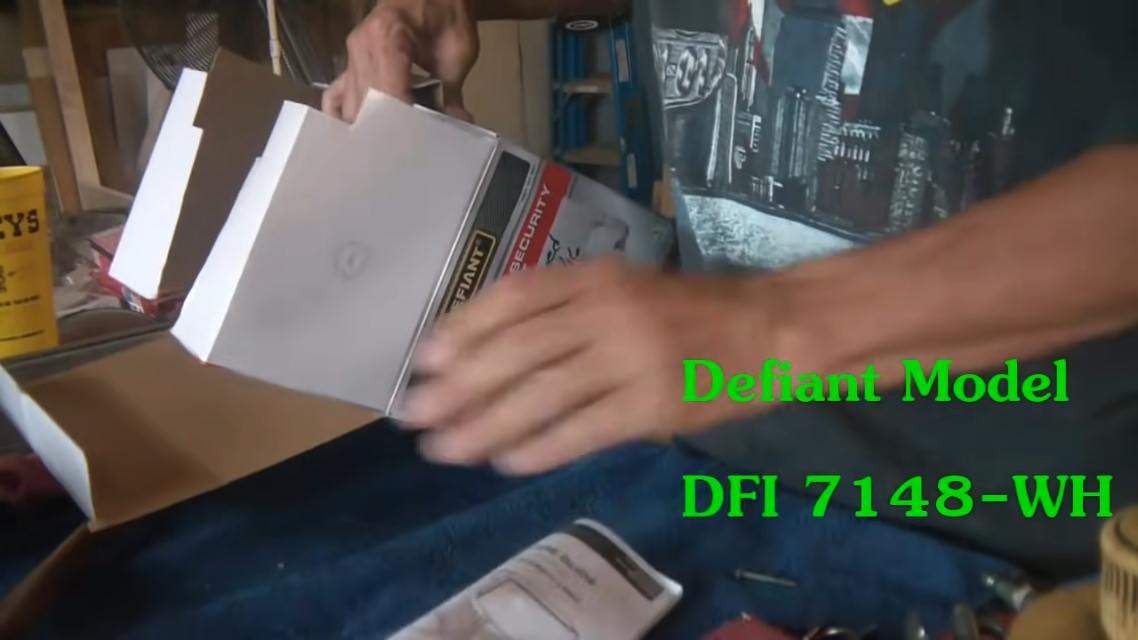}} & { just get to some box, check it out, see what all there is.} & {no model number of the product only the name of manufacturer of the product your reviewing.} \\ 
\hline

\multirow{2}{*}{\parbox{2cm}{\vspace{1.2cm} \centering Script\\Answerable}}  & {\includegraphics[width=0.25\textwidth]{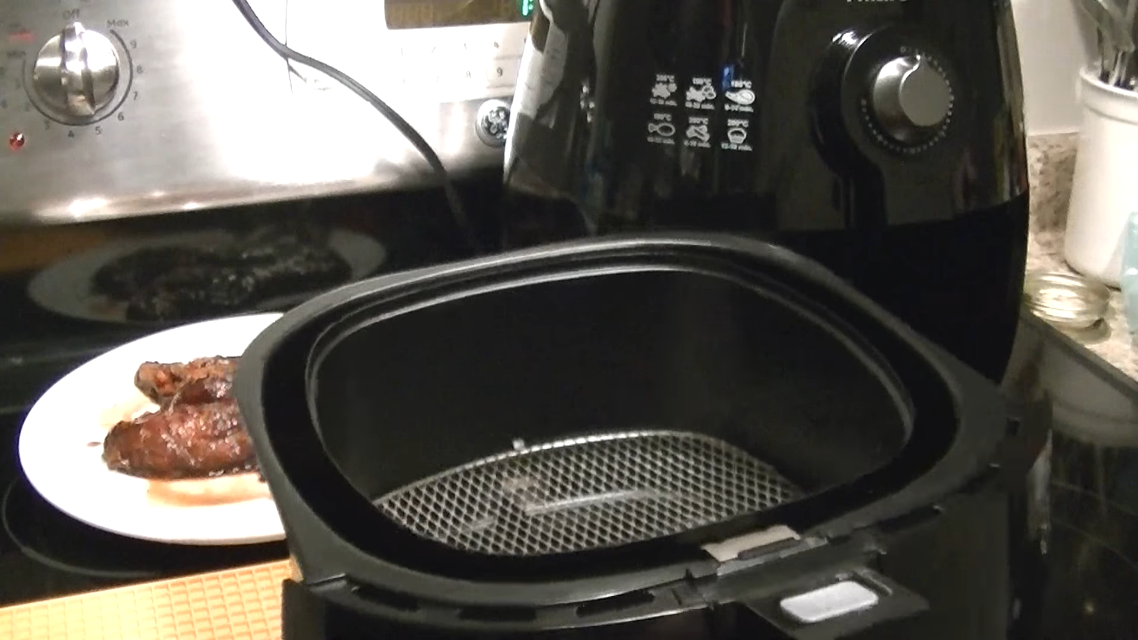}} & {right now I'm going to put it, already pre hit, at the air fryer for five minutes at 200 degrees Celsius.} & {what temperature did you cook it at?} \\ \cline{2-4} 
& {\includegraphics[width=0.25\textwidth]{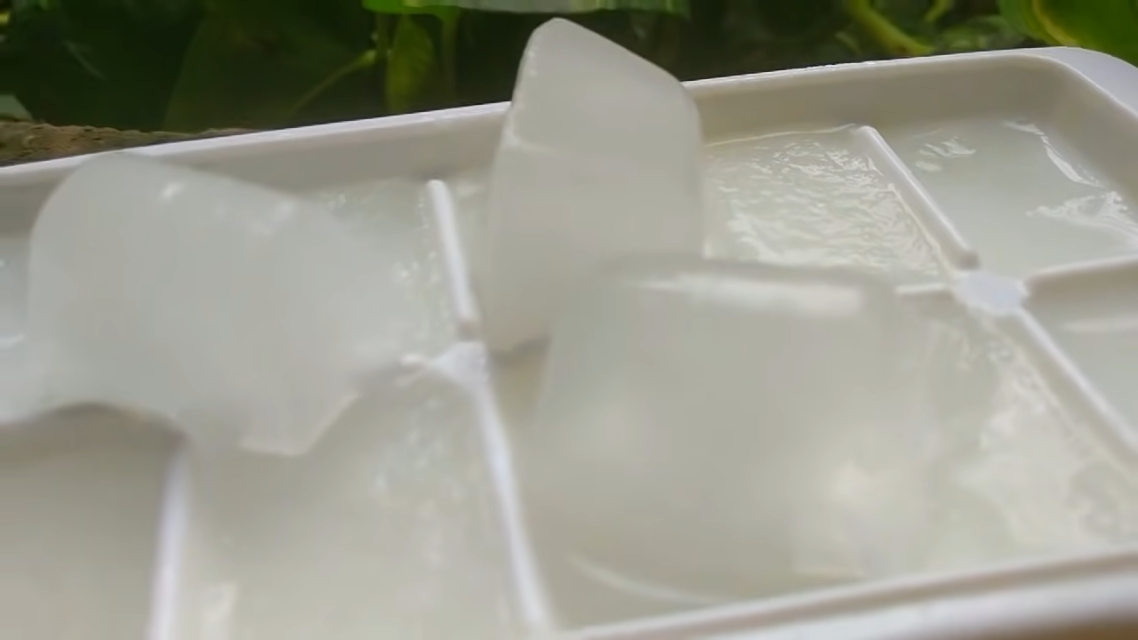}} & {rice water can be used on all skin types, such as dry, oily or normal skin.} & {mam can we use dis for dry skin} \\ 
\hline

\multirow{2}{*}{\parbox{2cm}{\vspace{1.2cm} \centering Combined\\Answerable}}  & {\includegraphics[width=0.25\textwidth]{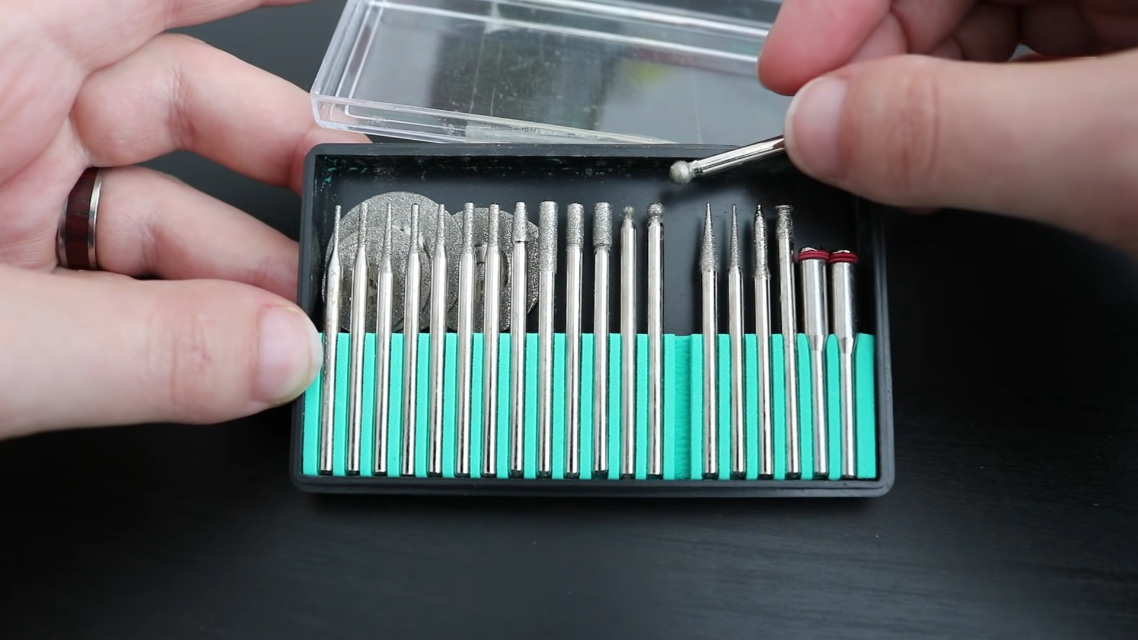}} & {There are a lot of different bits in this set, but a good place to start is with a ball shaped one.} & {Amazing! Can you tell me what bit did you use in this particular project?} \\ \cline{2-4} 
& {\includegraphics[width=0.25\textwidth]{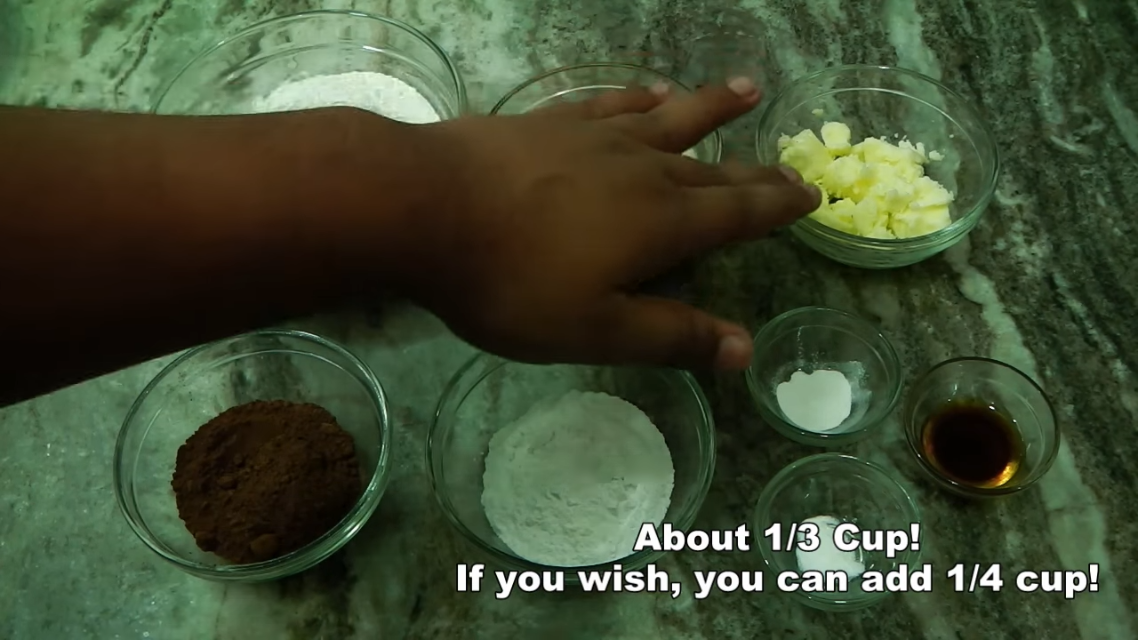}} & {first of all, one cup of maida, half a cup of condensed milk, little bit more than one foot cup of butter.} & {can u pls give measurement of ingredients also} \\ 

\bottomrule
\end{tabular}
\caption{Example questions of Visual answerable, Script answerable, and Combined answerable.}
\label{table:examples}
\end{table*}

As our questions naturally stem from real-world users rather than being crafted by hired workers, several distinctive traits emerge in our question set. First of all, our questions are roughly twice as long as those in existing Video QA datasets
\footnote{Ours: 17.2 vs. How2QA: $\sim$11, TutorialVQA: 9, HowToVQA69M: 8.7, iVQA: 7.6 words on average.}%
.
We believe this is partly because users often include context before or after their question to aid readers (i.e., content creators or other viewers) in better understanding the query. 

Another notable feature is their diversity, illustrated in Figure~\ref{figure:sunburst} by the initial two words of each question. Notably, the most common starting word, ``What", only consists of 13\% of all questions, in contrast to other Video QA datasets where ``What" is the clearly dominant initial word ~\cite{castro-etal-2022-in-the-wild, zhao2021photoshop, Li2020HeroHE}. Additionally, some questions start with casual words like ``hi" or ``great," reflecting users introducing their queries with greetings or expressions of appreciation in real-world situations. Some questions also include less polished or grammatically incorrect phrasing.
We show that there is minimal bias in answerability to specific question types in Appendix~\ref{sec:bias}.


\subsection{Analysis of Answerability Types}





Table~\ref{table:stats} presents the distribution of answerability types. Unanswerable questions make up 13.98\% of the total, while the remaining 86.02\% fall into the answerable category, encompassing different modalities required to answer: Visual, Script, and Combined (which requires both visual and script input). Below, we describe the characteristics of each answerability type. Examples of questions can be found in Table~\ref{table:examples}.

\subsubsection{(1) Unanswerable}
Approximately 13.98\% of the questions are deemed unanswerable, both in terms of visual and script cues. Although timestamps are used in replies of the questions, they serve more to clarify the question rather than pinpoint the answer segment (e.g., ``if you're talking about the tool at \texttt{timestamp}, it is ..."). Alternatively, replies contain relevant information related to the question, yet without providing a complete answer (e.g., ``I talk about X at \texttt{timestamp} if that helps any.")

\subsubsection{(2) Visual Answerable}
Approximately 61.19\% of questions find answers through visual snippets from the video, without requiring verbal input. Examples include queries about visually demonstrable tasks (e.g., ``how to take the top part off"). Another noteworthy instance occurs when visuals incorporate textual elements, such as names of depicted products. The practice of textual annotation is common in instructional videos, where it provides clarifications, corrections, or supplementary information that are not shown in the demonstration~\cite{chi2013democut}.

\subsubsection{(3) Script Answerable}
Roughly 49.53\% of questions find answers within the script. As the demonstrator visually guides through a process, they concurrently offer verbal explanations. Questions that seek specific details (e.g., ``what temperature was used for cooking?") are answered as the demonstrator provides explanations on them in the video.

Note that (2) Visual Answerable and (3) Script Answerable share overlapping questions. This arises from instances where a question can be answered solely through visuals \textit{and} solely through the script. Such cases include instances where the visual information closely mirrors the verbal content (or vice versa), such as when the creator verbally explains precisely what is occurring visually. It also includes cases where what the creator says are appear on visual through text. Approximately 32.16\% of questions are answerable by both modalities.

\subsubsection{(4) Combined Answerable}
Lastly, about 7.46\% of questions necessitate both visual and script inputs to provide answers, as these modalities complement each other. This encompasses scenarios like resolving references, where the author uses verbal cues such as ``here" or ``this" alongside visual demonstrations showing what they are. It also includes instances where visuals solidify verbal explanations by offering concrete examples (e.g., ``ball-shaped one"). Additionally, it includes cases where additional information is given in visuals through text annotations that complement what the author explains verbally.
\section{Experiments}
We benchmark the \sysname{} across two tasks regarding question answerability: (1) Segment Answerability Classification and (2) Video Answerability Classification. 
Below we describe the task, experiment setting, and results.

\subsection{Segment Answerability Classification}

\begin{table}[t]
\centering
\setlength{\tabcolsep}{20pt}
\begin{tabular}{ccc}
\toprule
\textbf{Model} & \textbf{Token} & \textbf{F1 score} \\ \midrule
\multicolumn{3}{c}{\textbf{\textit{Finetuned LMs}}} \\ 

{Llama-2 (7B)} & 4K & {45.78} \\ %
{Llama-2 (13B)} & 4K & {55.49} \\ %
\midrule
\multicolumn{3}{c}{\textbf{\textit{Zero-shot LMs}}} \\ 
{GPT-3.5} & 16K & {31.16} \\
{GPT-4} & 8K & {33.02} \\ %
\midrule
\multicolumn{3}{c}{\textbf{\textit{Multimodal Model}}} \\ 
{SeViLA} & 768 & {46.55} \\

\bottomrule
\end{tabular}
\caption{The results of Segmentation Answerability Classification task on \sysname{}. SeViLA selects 4 keyframes from 32 frames.}
\label{table:segment_results}
\end{table}

\subsubsection{Task Description}
Segment Answerability Classification asks the model to predict the answerability of questions given the evidence segment. Our evaluation methodology involves binary classification, where we categorize the answerability into 1) Unanswerable and 2) Answerable.

\subsubsection{Experimental Setup}
We conducted two types of experiments: 1) employing language models by textualizing both visual and verbal content of video segments~\cite{2023videochat} and 2) using multimodal models. For 1), we transcribed the visual content using an off-the-shelf image captioning model~\cite{vit-captioning} and used transcripts for the verbal content. Additionally, we observed that some questions that are Visual Answerable require the text in the visual content, such as text annotations detailing a product name (see Table~\ref{table:examples} for examples). Thus, we extracted text from the answer segment using the Tesseract OCR~\cite{TesseractOCR} and included it in the visual descriptions. Any text displayed within the timeframe was extracted and combined, and the leading author refined the OCR results. 

\subsubsection{Evaluation Metric}
We used F1 score for the evaluation metric. Responses that do not follow the prompt (i.e., that are neither ``Answerable" nor ``Unanswerable") have all been marked as incorrect.

\subsubsection{Baseline Models and Training Details}
We used both fine-tuning and zero-shot models for the language-based baseline models. We employed Llama2~\cite{Touvron2023Llama2O} for fine-tuning, which has demonstrated robust performance. We experimented with model sizes of 7B and 13B.
As for the zero-shot models, we utilized two powerful LLMs: ChatGPT~\cite{openai2023chatgpt} and GPT-4~\cite{openai2023gpt4}.
For the multimodal baseline, we used SeViLA~\cite{yu2023self}, which has demonstrated strong performance on video-language benchmarks.

We divided the collected data into training and evaluation sets in a 1:1 ratio to fine-tune the models. We addressed the class imbalance by augmenting unanswerable questions. This was achieved by integrating questions with descriptions of videos where the question had not been asked on. 
The prompts used for fine-tuning are outlined in Appendix~\ref{sec:prompts_segment} and training details in Appendix~\ref{sec:environment}.






\subsubsection{Results}
The experimental results are presented in Table~\ref{table:segment_results}. The results demonstrate the challenge of determining answerability. In particular, GPT-4 exhibited a tendency to misclassify answerable cases as unanswerable, mainly when dealing with visual information. A significant portion (53\%) of the false cases pertained to instances where GPT-4 mistakenly categorized Visual Answerable as Unanswerable.
The result with the multimodal baseline closely resembles that of Llama-2 (7B), which highlights the challenging nature of the problem even with a multimodal model.

\subsection{Video Answerability Classification}
\subsubsection{Task Description}
Video Answerability Classification task asks the model to predict the answerability of questions along with the required modality given the entire video. Our evaluation methodology involves multiple-choice question answering, where we categorize the answerability into five classes: 1) Unanswerable, 2) Visual Answerable, 3) Script Answerable, 4) Both Script and Visual Answerable, and 5) Combined Answerable.


\subsubsection{Experimental Setup}
Similar to the Segment Answerability Classification task, 
we conducted experiments with both 1) language and 2) multimodal baselines. For 1),
we first textualized both visual and verbal content of videos. However, as entire videos are often lengthy (8 min 44 sec on average), presenting the entire transcript, video caption and OCR text as input would exceed the model's capacity for training and inference. Therefore, we divided the video into segments corresponding to five transcript sentences. We then generated summaries for each segment using ChatGPT. For the verbal content, we provided the transcript, while for the visual content, we provided captions and OCR text. These verbal and visual summaries from each segment were then concatenated to create comprehensive visual and verbal descriptions for the entire video. 
More details about the prompts used for generating the summaries can be found in Appendix~\ref{sec:summary}.

\subsubsection{Evaluation Metric}
We used accuracy for the evaluation metric. Responses that do not follow the prompt (i.e., that are not classified as one of the five classes) have all been marked as incorrect.

\subsubsection{Baseline Models and Training Details}
We used the same baseline models as in the Segment Answerability Classification task: Llama2~\cite{Touvron2023Llama2O}, ChatGPT~\cite{openai2023chatgpt}, GPT-4~\cite{openai2023gpt4}, and SeviLa~\cite{yu2023self}.
However, we extended its context window for Llama2 by using rotary positional embeddings~\cite{su2022roformer} and the FlashAttention algorithm~\cite{dao2022flashattention} to incorporate the longer text inputs.
For SeViLA, we processed 768 tokens from the transcript, truncating sequences that exceeded the limit.

We divided the collected data into training and evaluation sets in a 1:1 ratio for fine-tuning. We oversampled Unanswerable, Visual Answerable, and Combined Answerable classes by a factor of two to address the class imbalance in the training set. 
The prompts used for fine-tuning are outlined in Appendix~\ref{sec:prompts} and training details in Appendix~\ref{sec:environment}.


\subsubsection{Results}

Table~\ref{table:answerable_results} shows the experimental results. The results highlight the difficulty in understanding extensive inputs encompassing video content and classifying the answerability with required modalities. 
In particular, GPT-4's false predictions predominantly leaned towards either Unanswerable or Script Answerable, indicating its heavy reliance on script information. Furthermore, a majority (85\%) of Combined Answerable instances were inaccurately predicted, indicating the complexities associated with processing multimodal inputs.
The result with the multimodal baseline falls slightly short of that achieved by Llama-2, highlighting the difficulty of the problem.

\begin{table}[t]
\centering
\setlength{\tabcolsep}{17pt}
\begin{tabular}{ccc}
\toprule
\textbf{Model} & \textbf{Token} & \textbf{Accuracy} \\ \midrule
\multicolumn{3}{c}{\textbf{\textit{Finetuned LMs}}}\\ 
{Llama-2 (7B)} & {16K} & {36.24} \\
{Llama-2 (13B)} & {16K} & {37.70} \\ \midrule
\multicolumn{3}{c}{\textbf{\textit{Zero-shot LMs}}} \\ 
{GPT-3.5} & {16K} & {18.42} \\ 
{GPT-4} & {8K} & {27.03} \\
\midrule
\multicolumn{3}{c}{\textbf{\textit{Multimodal Model}}} \\ 
{SeViLA} & 768 & {35.27} \\
\bottomrule
\end{tabular}
\caption{The results of Video Answerability Classification task on \sysname{}. SeViLA selects 4 keyframes from 32 frames.}
\label{table:answerable_results}
\end{table}
\section{Conclusion}
This paper presents the \sysname{} dataset, which contains questions generated by real-world users and the answerability information containing the required modality to provide answers. Through experiments with two answerability classification tasks, we illustrate the inherent challenges in discerning the answerability of video questions and the required modality for the answer. We belive that our work will foster the development of Video QA systems capable of identifying answer sources, thereby enhancing answer reliability. Furthermore, we hope that our work will encourage research that necessitates a deeper understanding and reasoning of multimodality in videos, where the verbal and visual aspects complement each other.

While our study underscores the significance of video question answerability, it has a few limitations. Our annotations primarily focused on the segments centered around the provided timestamps, which could have led to an incomplete understanding of the entire video context. Future research can investigate scenarios where insights from different segments, external knowledge, or commonsense affect answerability. 
Moreover, our experimental approach involved utilizing summaries or partial chunks of the video due to input length constraints.
This method might have overlooked specific video details. 
Subsequent efforts could explore incorporating complete and comprehensive video information to enhance long-form video question answerability.

\bibliography{aaai24}

\newpage
\appendix

\onecolumn 
\renewcommand{\thetable}{\Alph{table}}
\renewcommand{\thefigure}{\Alph{figure}}
\renewcommand{\thesection}{\Alph{section}}
\renewcommand{\theequation}{\Alph{equation}}
\renewcommand{\thealgorithm}{\Alph{algorithm}}
\renewcommand{\thesubsection}{\thesection.\alph{subsection}}
\setcounter{figure}{0}    
\setcounter{table}{0}    
\setcounter{equation}{0}
\setcounter{algorithm}{0}
\setcounter{footnote}{0} 
\setcounter{page}{1}
\section*{Appendix}

\section{Question Distribution Comparison} \label{sec:compare}

\begin{table}[ht]
\centering
\setlength{\tabcolsep}{8pt}
\begin{tabular}{lllllllllll}
\toprule
\textbf{YTCommentQA} & What & How & I & Where & Did & Can & Is & Hi & Why & Do \\
 Proportion (\%), N=2,332 & 12.99 & 8.79 &  6.65 & 5.02 & 4.8 & 4.46 & 3.34 & 3.34 & 3.13 & 2.4 \\ \midrule
\textbf{General questions }  &  What & I & How & Can & Is & Why & Who & Do & Hi & Where \\
Proportion(\%), N=3,736,077 & 9.16 & 7.23 & 6.41 & 6.26 & 4.26 & 4.26 & 2.82 & 2.65 & 2.64 & 2.56 \\
\bottomrule
\end{tabular}
\caption{Comparison between the distribution of the initial word in questions from \sysname{} and general questions, which were collected before applying the filtering criteria that require questions to contain at least one reply with a timestamp.}
\label{supp_table:stats}
\end{table}

\section{Prompts Used in Segment Answerability Classification Fine-Turning} \label{sec:prompts_segment}

\subsection{Language Model}
\begin{lstlisting}[breaklines=true, breakatwhitespace=true]
A chat between a curious human and an artificial intelligence assistant. The assistant gives helpful, detailed, and polite answers to the user's questions.

Please answer whether you can answer the given question based on the video segment description or not. Please provide your answer as either "Answerable" or "Unanswerable".

USER: Question: {Question}
Context: 
Transcript information: {Transcript information}
Visual information: {Visual information}

ASSISTANT:

\end{lstlisting}

\subsection{Multimodal Model}
\begin{lstlisting}[breaklines=true, breakatwhitespace=true]
{Video Frames} 
Question: {Question}
Context: {Transcript + OCR information}
Option A: The question cannot be answered from the given video.
Option B: The question can be answered from the given video.
Considering the information presented in the frame, select the correct answer from the options.
\end{lstlisting}

\section{Prompts Used in Video Answerability Classification Fine-Turning} \label{sec:prompts}
\subsection{Language Model}
\begin{lstlisting}[breaklines=true, breakatwhitespace=true]
A chat between a curious human and an artificial intelligence assistant. The assistant gives helpful, detailed, and polite answers to the user's questions.
We will provide you with the transcript and visual information of a video. As an AI assistant, you need to determine the question's answerability based on the information provided. Select the most appropriate option from below.
0: The question cannot be answered from the given video. 1: The question can be answered by the visual information, but cannot be answered by the transcript information. 2: The question can be answered by the transcript information, but cannot be answered by the visual information. 3: The question can be answered by the visual information alone. It can also be answered by the transcript information alone. 4: Both the transcript and visual information are required to answer the question. 
When you say the answer, you must say the number X only and X must be one of 0,1,2,3,4.

USER: Question: {Question}
Context: 
Transcript information: {Transcript information}
Visual information: {Visual information}

ASSISTANT:
\end{lstlisting}

\subsection{Multimodal Model}
\begin{lstlisting}[breaklines=true, breakatwhitespace=true]
{Video Frames} 
Question: {Question}
Context: {Transcript + OCR information}
Option A: The question cannot be answered from the given video.
Option B: The question cannot be answered by the script, but can be answered by the visual description.
Option C: The question can be answered with the script, but not with the visual description.
Option D: The question can be answered by the script only and also by the visual description only.
Option E: Both the script and visual description are required to answer the question.
Considering the information presented in the frame, select the correct answer from the options.
\end{lstlisting}

\section{Prompts Used in Video Summary Generation} \label{sec:summary}
\begin{lstlisting}[breaklines=true, breakatwhitespace=true]
Instruction:
A chat between a curious human and an artificial intelligence assistant. The assistant gives helpful, detailed, and polite answers to the user's questions.

Input:
Can you provide a short, comprehensive summary of the given text?
Context: {Transcript + Visual information}

Response: 
\end{lstlisting}

\section{Training Details} \label{sec:environment}
We trained the models until 3 epochs with a learning rate of 1e-5. All the experiments were conducted using 16 A100 GPUs.

\section{Bias in Answerability}\label{sec:bias}
To see if there is any bias in answerability depending on question types, we conducted additional experiments that predict answerability based on the frequency distribution of the first two words in our dataset. In our experiments on the same test dataset, 42.4\% of the test questions had no matching first two words with those in the training set, which demonstrates the diversity of our question set. For the remaining questions that had matching first two words, the results were as follows: An F1-score of 22.98 for the 1) Segment Answerability Classification task, which was lower than both language and multimodal baseline results. For 2) Video Answerability Classification task, an accuracy of 29.69 was obtained, which was lower than the fine-tuned language model and multimodal baseline results. These outcomes suggest that there is minimal bias in answerability that can be attributed to specific question types.

\section{Ethical Considerations}\label{sec:ethics}
We crawled YouTube comments only from publicly available videos. User names in the comments were anonymized with numbers (e.g., \texttt{User1}) in the annotation phase. Annotators provided consent to collect their Prolific IDs and responses.
Furthermore, we collected YouTube comments that must have gone through filtering of harmful content based on their guidelines. On top of this, the authors reviewed the comments to ensure that there was no harmful content in them.

\section{Snippets Shown Around Timestamp} \label{around_timestamp}

To extract the relevant visual and script snippets centered around the \texttt{timestamp}, we begin by identifying the transcript \texttt{(i)} where its start timestamp does not exceed the given \texttt{timestamp} with the minimum time gap.

\begin{figure}[ht]
\centering
\includegraphics[width=0.8\linewidth]{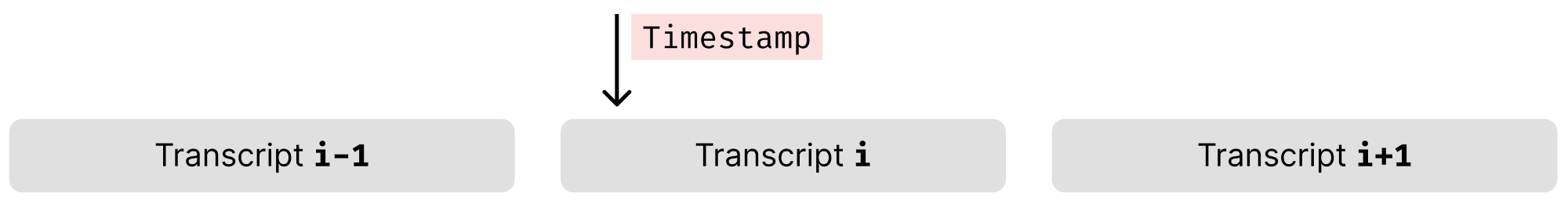}
\caption{We provide the video snippets centered around \texttt{timestamp}, aligned with the closest transcript and its corresponding visual content.
}
\label{figure:around}
\end{figure}

We then select the following transcripts: 
\begin{itemize}
    \item \textbf{Script}: preceding \texttt{(i-1)}, current \texttt{(i)}, and subsequent \texttt{(i+1)} transcripts
\end{itemize}

The selection of corresponding visual snippets is as follows:
\begin{itemize}
    \item \textbf{Visual}: 
    \begin{itemize}
        \item \textbf{Start}: \textit{max}((start timestamp of the first transcript + end timestamp of the previous transcript)/2, start timestamp of the first transcript - 5s))
        \item \textbf{End}: \textit{min}((end timestamp of the last transcript + start timestamp of the next transcript)/2, end timestamp of the last transcript + 5s)).
    \end{itemize}
\end{itemize}

However, if the corresponding visual snippets exceed 60 seconds, we chose the current \texttt{(i)} and the subsequent \texttt{(i+1)} transcripts. If this also surpasses 60 seconds, we selected the current \texttt{(i)} transcript only. If the duration continues to exceed 60 seconds, we did not include the question. 
Annotators were given the option to expand the script snippets by up to three additional sentences and their corresponding visual snippets.

\section{Answerability Annotation System} 
\label{sec:system}

\begin{figure}[ht]
\centering
\includegraphics[width=\linewidth]{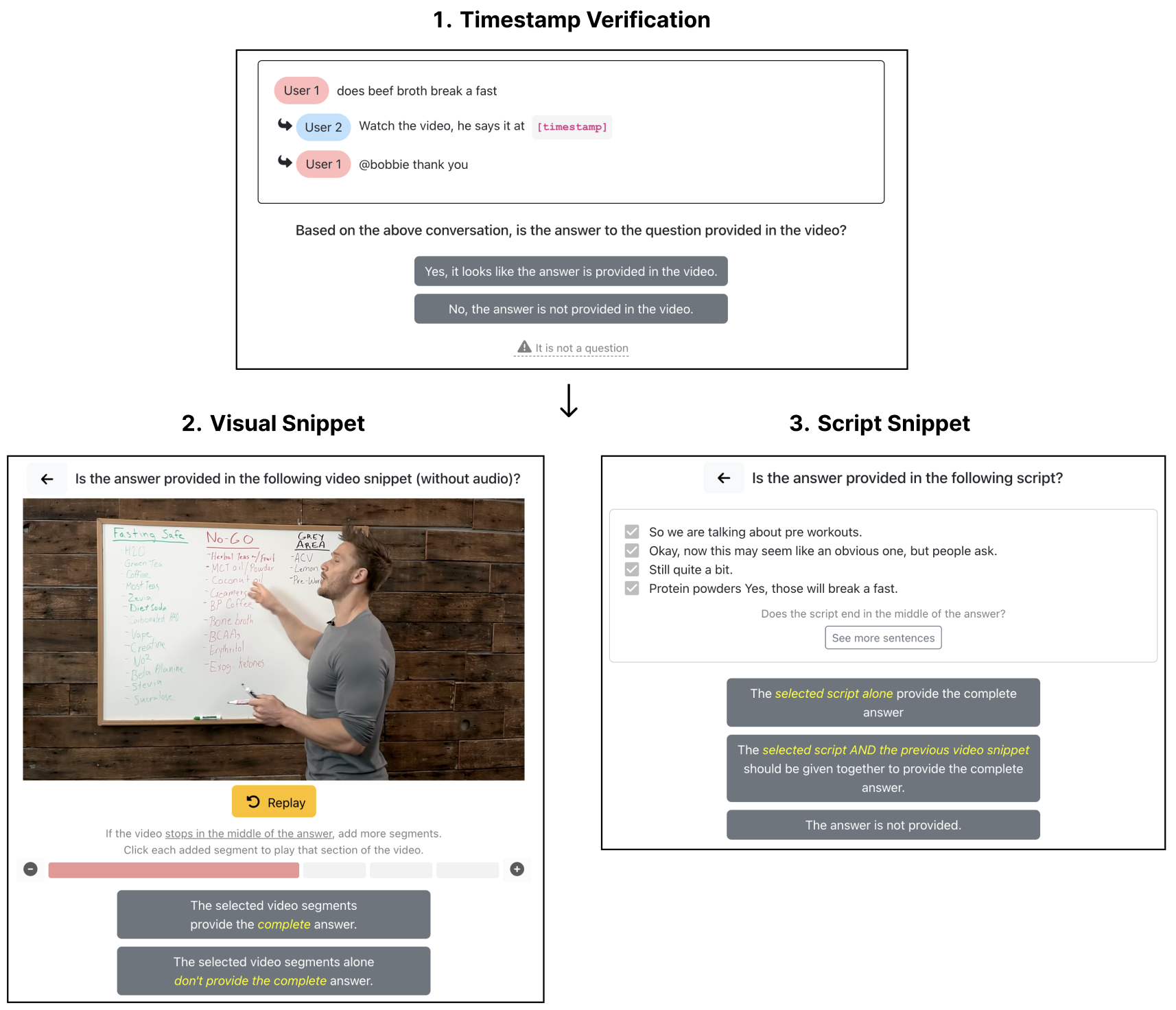}
\caption{System used in Video Answerability annotation. \textbf{1. Timestamp Verification}: Crowd workers were first presented with the entire conversation with the \texttt{timestamp} and asked if it implied that the answer to the question could be found around the video, or if the timestamp was used for a different purpose (e.g., timestamp of another video). Once they answered it did, they were moved on to the \textbf{2. Visual Snippet} phase where the corresponding video snippet without the audio was presented. They were asked if the question could be answered by the presented snippet or not. Then they were moved on to the \textbf{3. Script Snippet} phase where they were provided with the script snippet. They were asked if the question could be answered by the script, or if it required both the script and visual snippet, or if the answer was not provided. Note that \sysname{} includes questions that only passed the 1. Timestamp Verification phase.
}
\label{figure:system}
\end{figure}

\end{document}